\documentclass{article}
\usepackage{spconf,amsmath,graphicx}

\usepackage{url}
\usepackage{amsmath}
\usepackage{amssymb}
\usepackage{multirow}

\usepackage{graphicx}

\newcommand{\comment}[1]{}
\title{TreeGCN-ED: Encoding Point Cloud using a Tree-Structured Graph Network}

\name{Prajwal Singh \qquad Kaustubh Sadekar \qquad Shanmuganathan Raman}
  
\address{CVIG Lab, Indian Institute of Technology Gandhinagar \\ \{ singh\_prajwal, sadelkar.k, shanmuga\}@iitgn.ac.in}
\begin{document}
\ninept
\maketitle
\begin{abstract}

Point cloud is one of the widely used techniques for representing and storing 3D geometric data. In the past several methods have been proposed for processing point clouds. Methods such as PointNet and FoldingNet have shown promising results for tasks like 3D shape classification and segmentation. This work proposes a tree structured autoencoder framework to generate robust embeddings of point clouds by utilizing hierarchical information using graph convolution. We perform multiple experiments to assess the quality of embeddings generated by the proposed encoder architecture and visualize the t-SNE map to highlight its ability to distinguish between different object classes. We further demonstrate the applicability of the proposed framework in applications like: 3D point cloud completion and Single image based 3D reconstruction.

\end{abstract}
\section{Introduction}
\label{sec:intro}



Encoder-decoder based methods have been widely used for generating information preserving embeddings for different data modalities. In the past decades, several encoder-decoder based methods have been proposed for 2D images \cite{segnet, ronneberger2015unet, deeplab} and have shown impressive results for image compression and filtering tasks \cite{ffdnet, deepcnn}. However, it is challenging to extend these methods for 3D point cloud data due to its irregular structure as compared to that of images and 3D voxels. In this work, we focus on designing a deep-learning based encoder-decoder framework for processing point cloud data. 

Several methods have been proposed for encoding-decoding of point cloud data \cite{qi2017pointnet, pointcnn, voxelnet, pointrcnn}. PointNet \cite{qi2017pointnet} is a pioneer deep-learning-based method for encoding the point cloud data to lower-dimensional embeddings. These embeddings carry rich information about the point clouds and can be used for downstream tasks like segmentation and classification of point clouds. In a recent work \cite{shu20193d}, the authors propose a tree-structured decoder which uses the idea of graph convolution \cite{gcnn} to generate a point cloud using a noise vector $z \in R^{96}$ sampled from a normal distribution $\mathcal{N}(0, I)$. They propose aggregating the information from parent nodes at each layer instead of spatially adjacent nodes to leverage the tree-structured decoder architecture when applying graph convolution. The unique definition of graph convolution proposed in \cite{gcnn} highlights the effect of using information from parent nodes at multiple levels during the aggregation stage.

Inspired by the tree-structured decoder architecture in \cite{shu20193d}, we propose a tree-structured encoder and a graph convolution mechanism for down-sampling. \comment{which aggregates information from parent nodes instead of spatially neighbouring points similar to \cite{qi2017pointnet, yang2018foldingnet}}We combine our proposed encoder with the decoder proposed in \cite{shu20193d} to create a complete tree-based encoder-decoder framework called TreeGCN-ED \footnote{Code is available at: https://github.com/prajwalsingh/TreeGCN-ED} for processing point clouds. To show the effectiveness of proposed framework, we compare its results with FoldingNet \cite{yang2018foldingnet} architecture. The results show that TreeGCN-ED performs better than FoldingNet on two different evaluation metrics - Chamfer Distance (CD) \cite{fan2016point} and Fr\'echet point cloud distance (FPD) \cite{fan2016point}. We also observe that TreeGCN-ED learns inherent semantic information of the point cloud and hence, performs semantic segmentation without any explicit training. This highlights that our encoder generates more information-preserving embeddings. We also perform ablation studies to determine the effect of feature embedding dimension and data augmentation on the proposed TreeGCN-ED network. We further use the learned embeddings in a transfer learning setup and compare the results for point cloud classification on ModelNet10 and ModelNet40 datasets \cite{wu20153d}. Finally, we demonstrate the applicability of proposed framework for 3D point cloud completion and single image based 3D reconstruction.

\textbf{Contributions.} The following are the major contributions of this work.
\begin{itemize}
    \item A tree-structured encoder to generate robust embeddings for point cloud processing using graph convolution.
    \item An autoencoder based framework formed by the proposed encoder with the tree-based decoder \cite{shu20193d} for better point cloud reconstruction.
\end{itemize}

    \begin{figure*}[!htbp]
        \centering
        \includegraphics[width=1.0\textwidth]{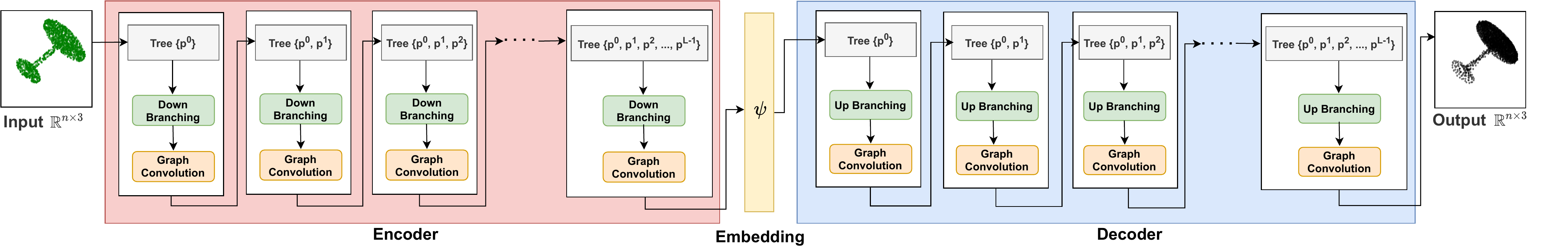}
        \caption{\textbf{TreeGCN-ED Architecture}. The encoder part (on the left) consist of downsampling and graph convolution modules for encoding the input 3D point cloud into a feature embedding $ \psi \in \mathbb{R}^{n}$. The decoder architecture (on right) is taking the embedding as input from the encoder and reconstructing the 3D point cloud.}
        \label{fig:treegcn_ed_model}
        \vspace{-3mm}
    \end{figure*}
    
    \begin{figure*}[!htbp]
        \centering
        \includegraphics[width=0.55\textwidth]{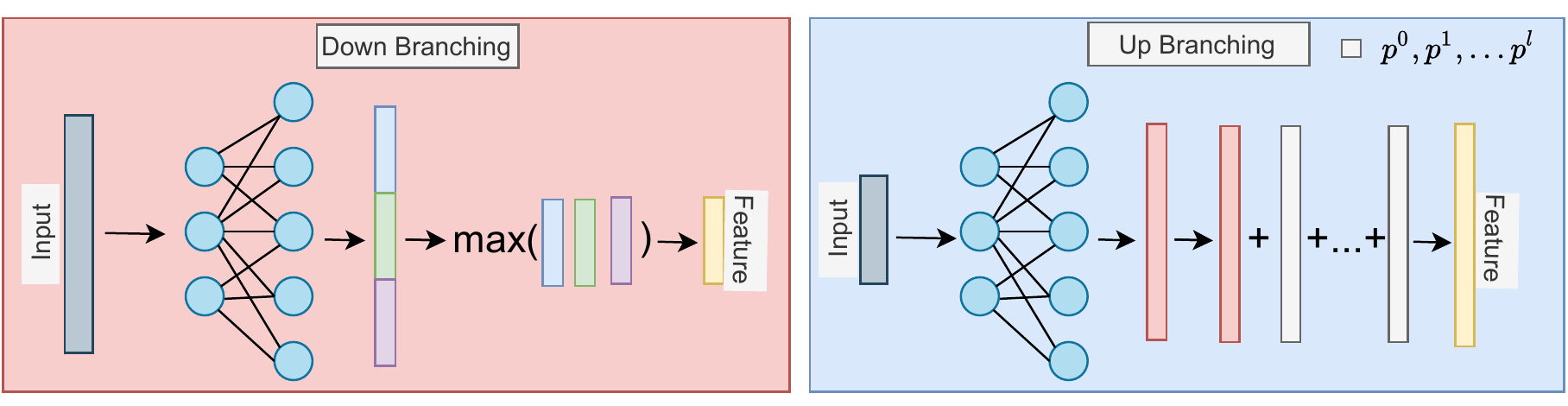}
        \caption{\textbf{TreeGCN-ED Down-Branching and Up-Branching Architecture}. The down-branching module (on the left) consists of a fully connected layer followed by max-pooling for feature extraction. Output of the fully connected layer is divided into $\mathbb{C}$ equal components that are passed to the max-pooling layer. The up-branching architecture is responsible for collecting information from the feature embedding of the ancestors and upsampling. The upsampled feature is passed to graph convolution layer for further processing.}
        \label{fig:treegcn_ed_submodel}
        \vspace{-3mm}
    \end{figure*}

\vspace{-2mm}
\section{RELATED WORK}
\label{sec:RelatedWork}

Point cloud is an important data structure that can be used to store information about the geometry of any 3D shape. There are various applications related to point cloud processing. Some of the applications and methods are point cloud classification and segmentation \cite{qi2017pointnet, UnstructuredPC, objectpart, Dohan2015LearningHS, Hackel2016FASTSS, Huang2016PointCL, dynamicgraphcnn}, point cloud completion \cite{yuan2019pcn, chang2015shapenet, Tchapmi_2019_CVPR, yang2018foldingnet}, point cloud auto-encoder \cite{yang2018foldingnet, qi2017pointnet, Kazhdan:2003:RIS, lfd, Girdhar16b, vconvdae, gan3d, lgan}, and Generative Adverserial Networks (GANs) for point cloud \cite{shu20193d, Li2019PointCG}.

In \cite{qi2017pointnet}, the authors propose the first end-to-end deep auto-encoder to directly process point cloud data. The encoder uses 1D CNN and global max-pooling to extract features of the input point cloud. This makes the model permutation invariant. The decoder reconstructs the point cloud using a three-layer fully connected network. FoldingNet \cite{yang2018foldingnet} model is build up on the idea of PointNet \cite{qi2017pointnet}, by proposing an auto-encoder network that uses graph-based method to learn encoding of a point cloud. Edge based convolution method is proposed in \cite{dynamicgraphcnn} to learn the local neighborhood as well as global properties of the 3D shape. In \cite{shu20193d}, the authors have proposed a deep generative model for 3D point cloud generation. This method is unique because the authors have used graph convolution on point cloud data which inherently does not contain any edge connections.


\section{METHOD}
\label{sec:method}

\subsection{Tree-GAN}
Tree-GAN \cite{shu20193d} proposes a deep generative model for 3D point cloud generation. It uses a branching method to gather information from neighbouring points. The accumulated information is then distributed to other points using graph convolution. The point cloud thus generated through this method is implicitly segmented.

In Tree-GAN \cite{shu20193d}, a noise vector $z \in \mathbb{R}^{96}$ is sampled from $\mathcal{N}(0, I)$ and is given as input to the generator network. Each layer of generator consists of a branching network and a graph convolution layer. The branching network accumulates the feature vectors from the previous layers which is then upsampled by the graph convolution layer to generate a new feature vector for that layer. This is repeated until the point cloud of the desired dimension $\mathbb{R}^{3 \times n}$ is obtained at the output. Note that the feature vector for the first layer is the noise vector $z$ itself. The generator and discriminator are trained under WGAN \cite{arjovsky2017wasserstein}.


\subsection{TreeGCN Based Point Cloud Encoder-Decoder}

\begin{figure*}[!htbp]
    \centering
    \includegraphics[width=1.0\textwidth]{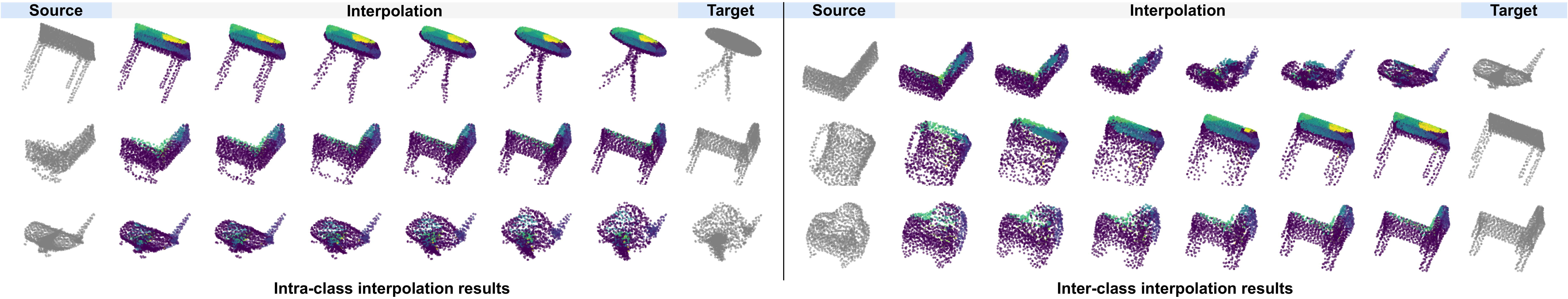}
    \caption{\textbf{Interpolation Result}. Illustration of intra-class (on left) and inter-class (on right) point cloud interpolation.}
    \label{fig:interpolation_result}
    \vspace{-3mm}
\end{figure*}


In this section, we discuss the proposed method for 3D point cloud processing. The key idea of this approach is inspired from the tree-GAN \cite{shu20193d}, a deep learning based model for generating a 3D point cloud from a noise vector. We use the idea of tree-based graph convolution from \cite{shu20193d} to develop an encoder that extracts rich embeddings to performs well on the unseen point cloud data. Our model takes a 3D point cloud of size $\mathbb{R}^{n \times 3}$ as input, then passes it through sequences of graph-based operation to generate encoding for the point cloud. The generated encoding is then passed through the decoder network where a sequence of graph-based operations upsample the encoding to obtain a $\mathbb{R}^{n \times 3}$ point cloud as the output. The complete network is trained end-to-end by minimizing chamfer loss \cite{fan2016point} and is called TreeGCN-ED.

Fig. \ref{fig:treegcn_ed_model} represents the proposed architecture of TreeGCN-ED. A $\mathbb{R}^{n \times 3}$ point cloud is given as input to the model, which then passes through a down branching network for gathering features from the ancestors of each node. Fig. \ref{fig:treegcn_ed_submodel} shows the down branching network and here, we are first passing each ancestor to a sequence of fully connected. Then, max pooling is applied on it to extract dominant features and we pass this feature to the next network, i.e. graph convolution network. The point cloud continuously passes through the down branching and graph convolution network sequence until the desired encoding $\psi$ is obtained. The generated encoding is given as input to the decoder network. In the decoder, the embedding is first passed through up branching network. The internal working of up branching network is shown in Fig. \ref{fig:treegcn_ed_submodel}, where the encoding is first passed through the fully connected layer. Then the feature vector is concatenated with the ancestor information, which can be useful for reconstructing point clouds. Then the constructed feature is passed on to the graph convolution network. This process is repeated till the point cloud of size $\mathbb{R}^{n \times 3}$ is reconstructed. The decoder architecture is similar to tree-GAN \cite{shu20193d}. The overall model is trained in an end-to-end manner using the chamfer loss function \cite{fan2016point}.

The branching network is an essential part of the TreeGCN-ED network. It helps in accumulating information from ancestors for each node. Every ancestor feature is passed through a fully connected layer at each stage to help the network learn the relation between a node and its neighbour. This is also useful because the point cloud does not have edge connections between points. We use max pooling for selecting important features from the encoded point cloud.  Max pooling has been proved to be a permutation invariant function \cite{qi2017pointnet}. We experimented with other pooling functions, such as averaging and adding feature vectors, but max-pooling works better than other methods. Tree graph convolution network learns the semantic segmentation of point cloud implicitly \cite{shu20193d}.

\subsection{Loss Function}

To train the TreeGCN-ED, we have used the chamfer loss \cite{fan2016point} function as it shows promising results for point cloud based reconstruction \cite{shu20193d}.
    
    \begin{equation}
    \resizebox{0.5\textwidth}{!}{
        \label{eqn:chamferloss}
        $
        \begin{aligned}
            \mathcal{L}_{chamfer}(S_{1}, S_{2}) &= \sum_{x \in S_{1}} min_{y \in S_{2}} || x - y ||_{2}^{2} + \sum_{y \in S_{2}} min_{x \in S_{1}} || x - y ||_{2}^{2}
        \end{aligned}$
        }
    \end{equation}
    
    In Equation \ref{eqn:chamferloss}, $S_{1} \in \mathbb{R}^{n \times 3}$ and $S_{2} \in \mathbb{R}^{n \times 3}$ represents two different point clouds. There are two specific reasons for using this loss function. First, it is permutation invariant \cite{fan2016point}. Second, it penalises the loss function if a point from one set is not matched with its corresponding nearest neighbour in another set and vice-versa. This forces the model to learn information preserving embedding for the point cloud.

\subsection{Data Preprocessing}

    To train our model, we have used ShapeNetBenchmarkV0 dataset \cite{chang2015shapenet} consisting of 16 object classes. The dataset is split into training, validation, and testing as per the standard ratio proposed in \cite{chang2015shapenet}. We uniformly sample $2048$ points from the meshes of the ShapeNet dataset \cite{chang2015shapenet}. To ensure uniform sampling of points, we make use of barycentric coordinates for the surface sampling.

\section{EXPERIMENTS AND RESULTS}
\label{sec:exp_eval}

\subsection{Training and Comparison of Encoder-Decoder Model}

We have used ShapeNetBenchmarkV0 dataset \cite{chang2015shapenet}, which consists of 16 object classes, with train-test split officially available along with the dataset. The dataset is first uniformly sampled for 2048 points and then passed on to the network. We have trained the complete network using the chamfer distance \cite{fan2016point} as the loss function till convergence. We compare the performance of TreeGCN-ED with the FoldingNet architecture \cite{yang2018foldingnet} on the test set of \cite{chang2015shapenet}. We use two different metrics for the evaluation task: Chamfer Distance (CD) \cite{fan2016point} and Fr\'echet Point Cloud Distance (FPD) \cite{shu20193d}. The results are shown in Table \ref{table:pointclouded} on ShapeNetBenchmarkV0 dataset \cite{chang2015shapenet}. For fair evaluation, in case of FoldingNet \cite{yang2018foldingnet}, we have considered the first $2025$ minimum point cloud distances for calculating CD, owing to the difference in input and output point cloud size. The quantitative results shown in the table highlight that our proposed method performs better than FoldingNet \cite{yang2018foldingnet} for chamfer distance (CD) as well as Fr\'echet point cloud distance (FPD).



\begin{table*}[!htbp]
\resizebox{\textwidth}{!}{
\begin{tabular}{clllllllllllllllll|l}
\hline
\multirow{2}{*}{Models} & \multirow{2}{*}{Metrics}                                             & \multicolumn{16}{c|}{Object Class}                                                                                                                                                                                                                                                                                                                                                                                                                                                                                                                                                                                                                                                                                                                                                                                                                                                                               &                                                               \\
                        &                                                                      & \multicolumn{1}{c}{Airplane}                         & \multicolumn{1}{c}{Bag}                              & \multicolumn{1}{c}{Cap}                               & \multicolumn{1}{c}{Car}                              & \multicolumn{1}{c}{Chair}                            & \multicolumn{1}{c}{Earphone}                          & \multicolumn{1}{c}{Guitar}                           & \multicolumn{1}{c}{Knife}                            & \multicolumn{1}{c}{Lamp}                             & \multicolumn{1}{c}{Laptop}                           & \multicolumn{1}{c}{Motorbike}                        & \multicolumn{1}{c}{Mug}                              & \multicolumn{1}{c}{Pistol}                           & \multicolumn{1}{c}{Rocket}                           & \multicolumn{1}{c}{Skateboard}                       & \multicolumn{1}{c|}{Table}                            & Average                                                         \\ \hline
FoldingNet \cite{yang2018foldingnet}             & \multicolumn{1}{c}{\begin{tabular}[c]{@{}c@{}}CD\\ FPD\end{tabular}} & \begin{tabular}[c]{@{}l@{}}0.67\\ 11.10\end{tabular} & \begin{tabular}[c]{@{}l@{}}3.12\\ 87.45\end{tabular} & \begin{tabular}[c]{@{}l@{}}2.82\\ 117.36\end{tabular} & \begin{tabular}[c]{@{}l@{}}1.76\\ 28.47\end{tabular} & \begin{tabular}[c]{@{}l@{}}1.47\\ 12.00\end{tabular} & \begin{tabular}[c]{@{}l@{}}3.34\\ 152.04\end{tabular} & \begin{tabular}[c]{@{}l@{}}0.44\\ 19.55\end{tabular} & \begin{tabular}[c]{@{}l@{}}0.55\\ 19.56\end{tabular} & \begin{tabular}[c]{@{}l@{}}2.60\\ 45.19\end{tabular} & \begin{tabular}[c]{@{}l@{}}1.01\\ 11.19\end{tabular} & \begin{tabular}[c]{@{}l@{}}1.48\\ 33.91\end{tabular} & \begin{tabular}[c]{@{}l@{}}2.28\\ 40.17\end{tabular} & \begin{tabular}[c]{@{}l@{}}1.16\\ 30.14\end{tabular} & \begin{tabular}[c]{@{}l@{}}0.88\\ 32.53\end{tabular} & \begin{tabular}[c]{@{}l@{}}1.35\\ 47.17\end{tabular} & \begin{tabular}[c]{@{}l@{}}1.70\\ 24.62\end{tabular} & \begin{tabular}[c]{@{}l@{}}1.48\\ 44.52\end{tabular}          \\ \hline
TreeGCN-ED              & \multicolumn{1}{c}{\begin{tabular}[c]{@{}c@{}}CD\\ FPD\end{tabular}} & \begin{tabular}[c]{@{}l@{}}0.50\\ 5.79\end{tabular}  & \begin{tabular}[c]{@{}l@{}}1.88\\ 21.02\end{tabular} & \begin{tabular}[c]{@{}l@{}}1.62\\ 16.14\end{tabular}  & \begin{tabular}[c]{@{}l@{}}1.45\\ 9.47\end{tabular}  & \begin{tabular}[c]{@{}l@{}}1.32\\ 7.85\end{tabular}  & \begin{tabular}[c]{@{}l@{}}1.91\\ 51.79\end{tabular}  & \begin{tabular}[c]{@{}l@{}}0.40\\ 13.90\end{tabular} & \begin{tabular}[c]{@{}l@{}}0.41\\ 14.80\end{tabular} & \begin{tabular}[c]{@{}l@{}}1.97\\ 21.82\end{tabular} & \begin{tabular}[c]{@{}l@{}}0.88\\ 2.56\end{tabular}  & \begin{tabular}[c]{@{}l@{}}1.14\\ 14.67\end{tabular} & \begin{tabular}[c]{@{}l@{}}1.72\\ 12.70\end{tabular} & \begin{tabular}[c]{@{}l@{}}0.79\\ 9.62\end{tabular}  & \begin{tabular}[c]{@{}l@{}}0.61\\ 23.91\end{tabular} & \begin{tabular}[c]{@{}l@{}}0.78\\ 13.90\end{tabular} & \begin{tabular}[c]{@{}l@{}}1.41\\ 13.90\end{tabular} & \textbf{\begin{tabular}[c]{@{}l@{}}1.21\\ 11.54\end{tabular}} \\ \hline
\end{tabular}
}
\caption{Comparison of the efficiency for 3D point cloud encoding-decoding between our proposed architecture and the FoldingNet \cite{yang2018foldingnet} model on ShapeNetBenchmarkV0 dataset \cite{chang2015shapenet}.}
\label{table:pointclouded}
\vspace{-2mm}
\end{table*}

\subsection{Point Cloud Interpolation}

To show that our proposed encoder architecture is learning information rich embedding, we perform inter-class and intra-class interpolation experiments between the source and the target point clouds. The interpolation results are shown in Fig. \ref{fig:interpolation_result}.

The intra-class interpolation results illustrate the ability of our model to synthesize novel shapes between two given shapes. We observe that the generated shapes faithfully represent the object class at each interpolation stage and the interpolation is observed to be very smooth. Similarly, in the case of inter-class interpolation, we observe a smooth transition of characteristic class features from one object class to another.

\subsection{t-SNE Visualization}

\begin{figure}[!htbp]
    \centering
    \includegraphics[width=0.5\textwidth]{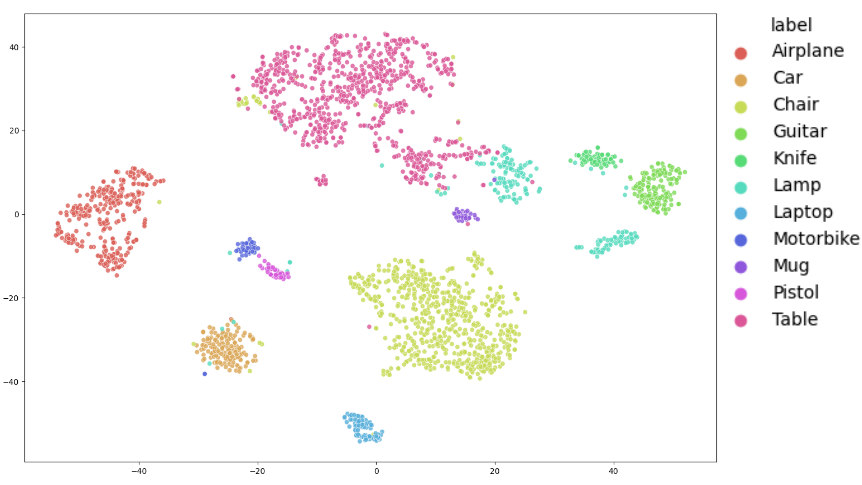}
    \caption{The visualization of t-SNE \cite{tsne} clustering of the feature embeddings obtained from TreeGCN-ED model.}
    \label{fig:embedding_space}
    \vspace{-5mm}
\end{figure}

We use t-SNE \cite{tsne} plot to show how well our encoder model can generate feature embedding for each class. We set the perplexity value to 40. Based on the results of t-SNE plot shown in Fig. \ref{fig:embedding_space}, the inter-class separation is higher. This signifies the discriminative capacity of our proposed encoder model.

\subsection{Ablation Studies}

We perform ablation studies to determine how feature embedding $\psi$ dimension and data augmentation affect the ability of TreeGCN-ED to learn a meaningful feature representation. Four different training regimes are compared on the ShapeNetCore.v2 test-set \cite{largeshapenet} which consist of 55 classes. In \textit{Regime 1 and 2}, the dimension of feature embedding is fixed to 256 and 512, respectively, without augmentation. Similarly, in \textit{Regime 3 and 4}, the dimension of feature embedding is fixed to 256 and 512, respectively, but with augmentation. We use ShapeNetCore.v2 dataset \cite{largeshapenet} to train TreeGCN-ED for all the four regimes. Table \ref{table:exp_cd} shows the variation in chamfer distance for the above mentioned regimes. Based on the results, Regime 4 gives the best model performance.

\begin{table}[!htbp]
\centering
\footnotesize
\begin{tabular}{lllll}
\hline
\multicolumn{1}{c}{\multirow{2}{*}{Dataset}} & \multicolumn{2}{c}{No Augmentation} & \multicolumn{2}{c}{Rotation Augmentation} \\
\multicolumn{1}{c}{}         & $\psi = 256$     & $\psi = 512$     & $\psi = 256$        & $\psi = 512$        \\ \hline
ShapeNetCore.v2 & 10.90            & 10.07            & 8.82                & 7.88                \\ \hline
\end{tabular}
\caption{Quantitative results for all four regimes for the task of 3D point cloud encoding-decoding. Chamfer distance \cite{fan2016point} is used as the metric for comparison.}
\label{table:exp_cd}
\vspace{-5mm}
\end{table}

Furthermore, we also evaluate the efficiency of feature representation learning of TreeGCN-ED on ModelNet10 and ModelNet40 datasets \cite{wu20153d} for all the four regimes. We follow the same procedure as mentioned in \cite{yang2018foldingnet} to train a linear SVM classifier on features extracted from trained TreeGCN-ED for the ModelNet datasets \cite{wu20153d}. Table \ref{table:exp_acc} shows the variation in classification accuracy for all the four regimes on the test set of ModelNet datasets \cite{wu20153d}. Based on the results, Regime 4 gives the best model performance. 

\begin{table}[!htbp]
\centering
\footnotesize
\begin{tabular}{lllll}
\hline
\multicolumn{1}{c}{\multirow{2}{*}{Dataset}}  & \multicolumn{2}{c}{No Augmentation} & \multicolumn{2}{c}{Rotation Augmentation} \\
\multicolumn{1}{c}{}    & $\psi = 256$     & $\psi = 512$     & $\psi = 256$        & $\psi = 512$        \\ \hline
ModelNet10 & 0.83             & 0.83             & 0.85                & 0.85                \\
ModelNet40 & 0.71             & 0.72             & 0.73                & 0.73                \\ \hline
\end{tabular}
\caption{Quantitative results for all four regimes for the task of 3D point cloud classification using transfer learning on ModelNet10 and ModelNet40 dataset \cite{wu20153d}.}
\label{table:exp_acc}
\vspace{-2mm}
\end{table}

It can be easily argued that the tree-GAN \cite{shu20193d} decoder itself is enough for point cloud processing at hand. However, to establish the need and examine the strength of the proposed encoder, we perform an additional experiment by replacing it with the PointNet \cite{qi2017pointnet} encoder to train the complete network on ShapeNetBenchmarkV0 dataset \cite{chang2015shapenet}. We observed that the average CD is $8.65$ with the PointNet encoder and $1.21$ with the proposed encoder. This clearly establishes the efficacy of the proposed encoder.

\subsection{Applications}

We showcase two potential applications of our proposed method: 3D point cloud completion and single image based 3D reconstruction. 

\subsubsection{3D Point Cloud Completion}

\begin{figure}[!htbp]
    \centering
    \includegraphics[width=0.5\textwidth]{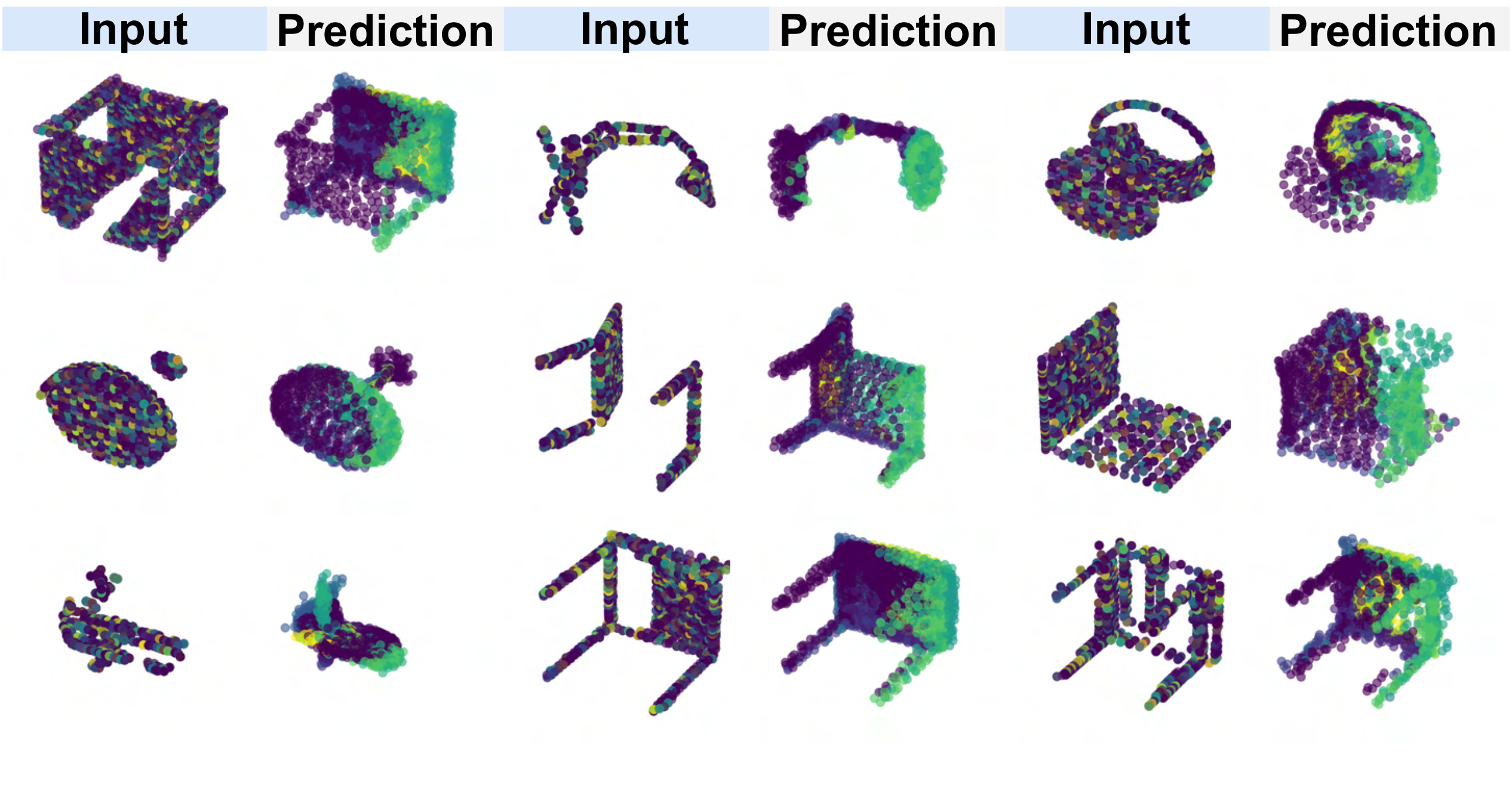}
    \caption{Qualitative results on the test set of Completion 3D benchmark dataset \cite{yuan2019pcn, chang2015shapenet, Tchapmi_2019_CVPR}.}
    \label{fig:shapecompletion}
    \vspace{-5mm}
\end{figure}

The task of 3D point cloud completion is to reconstruct the incomplete input point cloud. To achieve this, we train TreeGCN-ED on the Completion 3D benchmark dataset \cite{yuan2019pcn, chang2015shapenet, Tchapmi_2019_CVPR} and perform a qualitative evaluation on the officially available test set. The qualitative results are shown in Fig. \ref{fig:shapecompletion}. Since the ground truth for the test set is not available, we only perform qualitative analysis.

\subsubsection{Single Image Based 3D Reconstruction}

\begin{figure}[!htbp]
    \centering
    \includegraphics[width=0.5\textwidth]{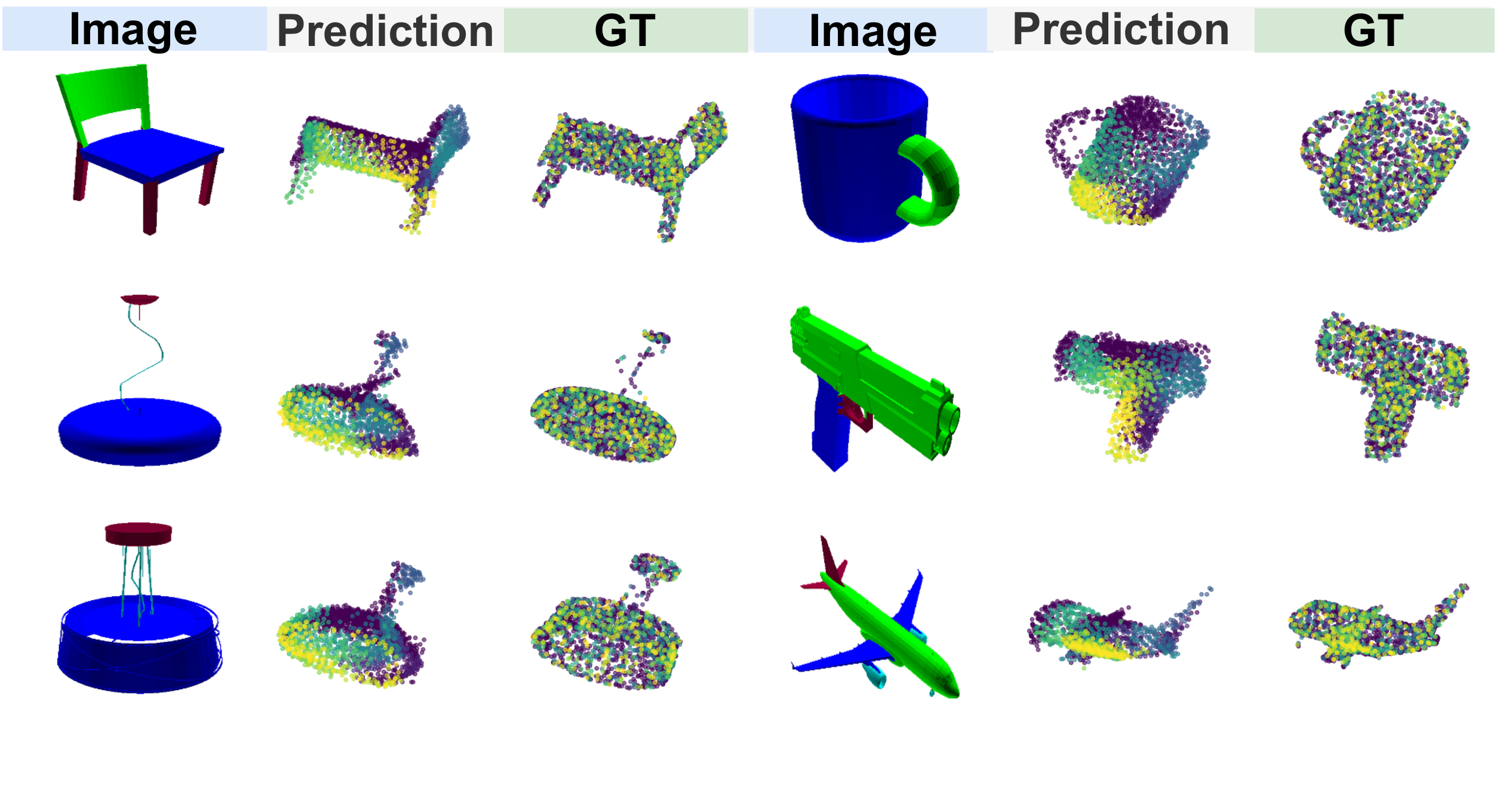}
    \caption{Qualitative results for single image to 3D reconstruction on the test set of ShapeNetBenchmarkV0 dataset \cite{chang2015shapenet}.}
    \label{fig:image23d}
    \vspace{-5mm}
\end{figure}

3D reconstruction from a single image is an ill-posed problem. This problem arises due to the ambiguity involved in the occluded part of the object, which is not visible in the image. We attempt to solve this problem using TreeGCN-ED architecture. We train the TreeGCN-ED model on 16 different classes of the ShapeNetBenchmarkV0 dataset \cite{chang2015shapenet} till convergence. Later, we replace the encoder of TreeGCN-ED with a CNN based architecture to extract image features. We freeze the trained weights of the decoder and train the image encoder network end-to-end for 3D reconstruction. We use Chamfer Distance (CD) \cite{fan2016point} as the loss function. We use the synthesized images available in ShapeNetBenchmarkV0 dataset \cite{chang2015shapenet} to train the single image to 3D shape reconstruction model. The qualitative results are shown in Fig. \ref{fig:image23d}.

\section{CONCLUSION}
\label{sec:conclusion}

In this work, we propose a tree-structured graph convolution-based encoder architecture and combine it with the decoder of tree-GAN to create a complete tree-structured encoder-decoder model for processing 3D point cloud data. The experimental results of our proposed architecture highlight the effectiveness of the encoder model in learning information-rich features. We also showcase that TreeGCN-ED can be used for the task of point cloud completion and single image based 3D reconstruction. 

\section{Acknowledgments}

This research is supported by SERB MATRICS and SERB IMPRINT-2 grants. Also, we would like to thank Ashish Tiwari and Dhananjay Singh for their constructive and valuable feedback.


\bibliographystyle{IEEEbib}
\small{
    \bibliography{strings,refs}
}

\end{document}